\documentclass{article}
\usepackage{ijcai26}

\usepackage{times}
\usepackage{soul}
\usepackage{url}
\usepackage[hidelinks]{hyperref}
\usepackage[utf8]{inputenc}
\usepackage[small]{caption}
\usepackage{graphicx}
\usepackage{amsmath}
\usepackage{amsthm}
\usepackage{booktabs}
\usepackage{algorithm}
\usepackage{algorithmic}
\usepackage{amsmath,amsfonts}
\usepackage{textcomp}
\usepackage{stfloats}
\usepackage{url}
\usepackage{verbatim}
\usepackage{graphicx}
\usepackage{tabularx}                  
\usepackage{booktabs}                  
\usepackage{multirow}                  
\newcolumntype{C}{>{\centering\arraybackslash}X} 
\newcolumntype{M}{>{\centering\arraybackslash}m{0.27\linewidth}}

\usepackage[switch]{lineno}

\title{Easy Adaptation: An Efficient Task-Specific Knowledge Injection Method for Large Models in Resource-Constrained Environments}

\author{
    Dong Chen, Zhengqing Hu, Shixing Zhao, Yibo Guo
}

\begin{document}

\maketitle

\begin{abstract}
While the enormous parameter scale endows Large Models (LMs) with unparalleled performance, it also limits their adaptability across specific tasks. Parameter-Efficient Fine-Tuning (PEFT) has emerged as a critical approach for effectively adapting LMs to a diverse range of downstream tasks. However, existing PEFT methods face two primary challenges: (1) High resource cost. Although PEFT methods significantly reduce resource demands compared to full fine-tuning, it still requires substantial time and memory, making it impractical in resource-constrained environments. (2) Parameter dependency. PEFT methods heavily rely on updating a subset of parameters associated with LMs to incorporate task-specific knowledge. Yet, due to increasing competition in the LMs landscape, many companies have adopted closed-source policies for their leading models, offering access only via Application Programming Interface (APIs). Whereas, the expense is often cost-prohibitive and difficult to sustain, as the fine-tuning process of LMs is extremely slow. Even if small models perform far worse than LMs in general, they can achieve superior results on particular distributions while requiring only minimal resources. Motivated by this insight, we propose Easy Adaptation (EA), which designs Specific Small Models (SSMs) to complement the underfitted data distribution for LMs. Extensive experiments show that EA matches the performance of PEFT on diverse tasks without accessing LM parameters, and requires only minimal resources.
\end{abstract}

\section{Introduction}

In recent years, Large Models (LMs) have achieved remarkable success \cite{zheng2025large}. However, as their parameter scale continue to grow, the computational resources required for fine-tuning have become increasingly burdensome. For instance, fine-tuning LLaMA-65B \cite{touvron2023llama} requires ten GPUs with 80 GB of memory each, making the process prohibitively costly for most individuals and thereby motivating a growing reliance on cloud-based LM services \cite{borzunov2023distributed,chen2025logic}.

To reduce the computational cost of fine-tuning and enable LMs to adapt to various specific tasks, Parameter-Efficient Fine-Tuning (PEFT) have been proposed, which train only a small subset of parameters while keeping the majority of parameters unchanged \cite{han2024parameter,lei2023conditional}. 
For example, for the most popular PEFT methods Low-Rank Adaptation (LoRA) \cite{hu2022lora}, in the case of fine-tuning GPT-3, LoRA can reduce hardware requirements by a factor of three compared to the full fine-tuning.
\begin{figure}[t]
	\centering
	\includegraphics[width=8.6cm,height=4.5cm]{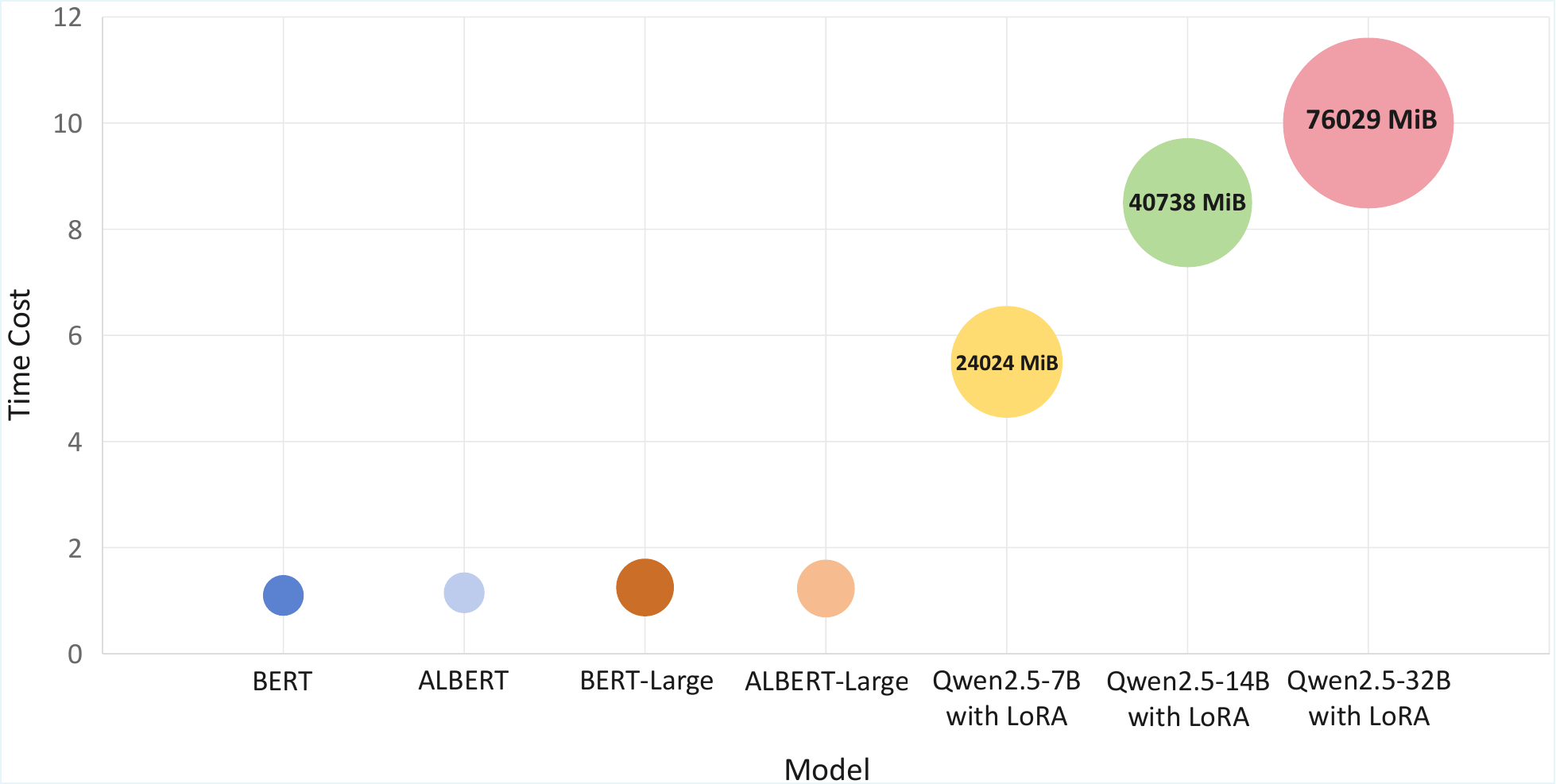}
	\caption{Cost of training. The vertical axis indicates the time cost, while the size of each bubble corresponds to the memory cost.}
	\label{Figure: memory and time}
\end{figure}

Although PEFT methods have demonstrated impressive performance, it remains challenging to adapt LMs to specific tasks, which can be summarized as: 
(1) High resource cost. As shown in Figure \ref{Figure: memory and time}, when fine-tuning with LoRA, the memory and time cost for fine-tuning a LM remain significantly higher than those for training a small model. For instance, fine-tuning Qwen2.5-14B with LoRA requires about 40 GB of memory, and the training time with the same setting is approximately nine times longer compared to that of a small model. High resource costs hinder the effectiveness of PEFT in resource-constrained environments, as the vast majority of individual devices still remain at the same level as before the era of LMs.
(2) Parameter dependency. As competition among LMs intensifies, many companies have chosen to close-source their leading LMs, providing only limited access via APIs to safeguard proprietary technologies. Fine-tuning these closed-source LMs in the cloud has become prohibitively expensive. For instance, training the o4-mini model via APIs incurs a cost of \$100 per hour \footnote{https://openai.com/api/pricing/}. Given that fine-tuning LMs is often a time-consuming process, this can result in significant expenses. Furthermore, existing PEFT methods heavily depend on the parameters update, rendering them unsuitable for fine-tuning the closed-source models.

Prior studies have demonstrated that models with larger parameter scales can accommodate a broader range of data distributions \cite{kaplan2020scaling,bahri2024explaining}. Consequently, LMs with extensive prior knowledge generally outperform Specific Small Models (SSMs) that are trained on specific datasets.
Although SSMs lag behind LMs in terms of overall performance, existing works indicate that SSMs can achieve comparable or even superior results to LMs on particular data distributions \cite{chen2024data,chen2025improving}. Therefore, we propose to complement the underfitted distributions of LMs by training SSMs, thereby mitigating the aforementioned limitations of PEFT methods.

\begin{figure*}[t]
	\centering
	\includegraphics[scale=0.4]{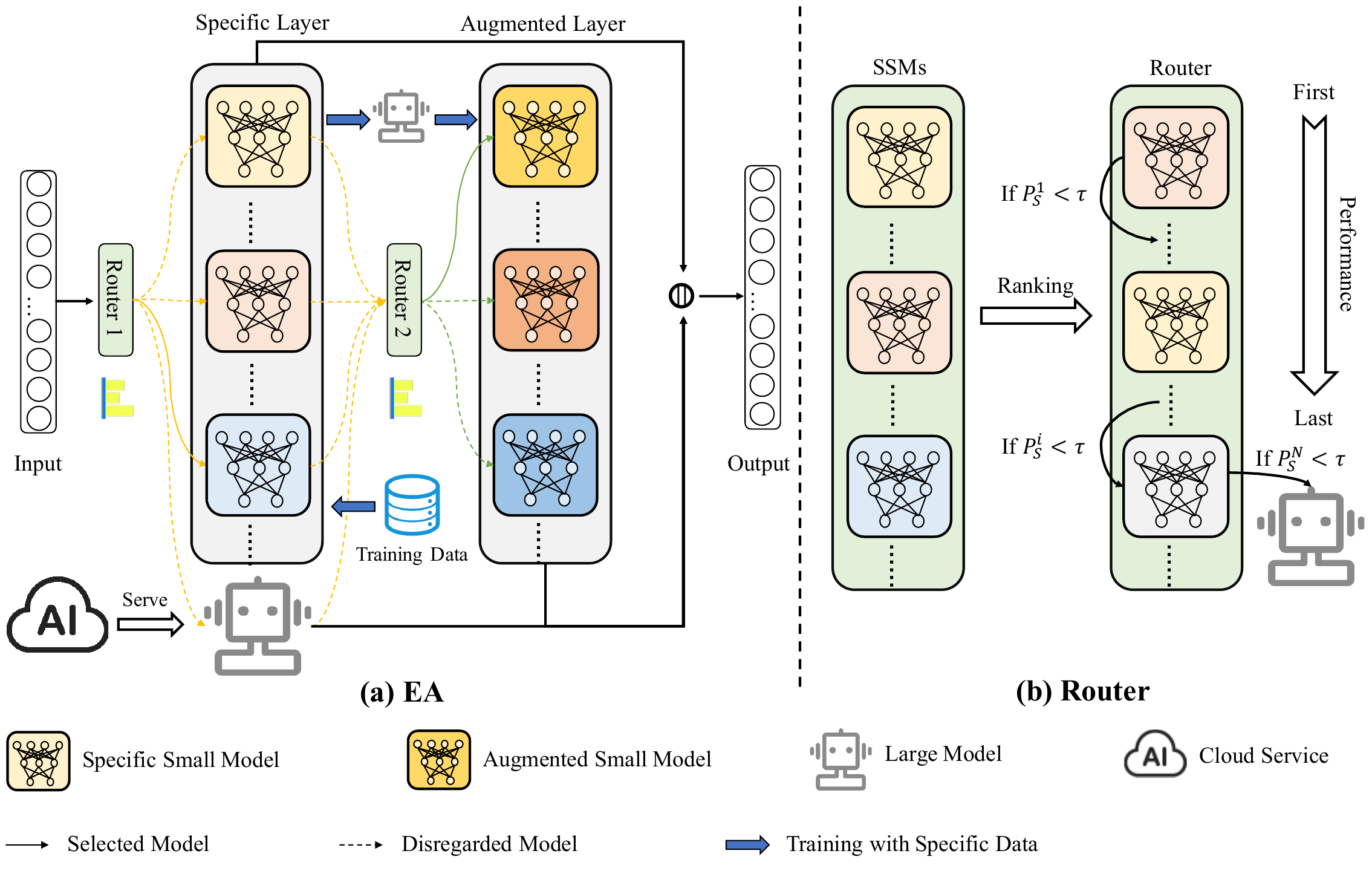}
	\caption{The framework of the proposed EA. }
	\label{Figure: CA pipline}
\end{figure*}

As illustrated in Figure \ref{Figure: CA pipline} (a), we propose Easy Adaptation (EA) based on the collaboration between cloud-based LMs and tailored SSMs, enabling the vast majority of individuals to efficiently inject task-specific knowledge into LMs even in resource-constrained environments. Specifically, when adapting a LM to a specific task, EA first trains SSMs to coarse-grained complement the underfitted data distribution of the LM and employs a Router to match input to the most appropriate model. Subsequently, EA selects underfitted training data with the existing SSMs and the LM, thereby targeting and compensating for the current capability deficiencies to further enhance overall performance.
Extensive experiments on various tasks demonstrate that EA can achieve performance comparable to or surpassing PEFT methods like LoRA, while significantly reducing training cost. Particularly,  in image classification tasks, EA enhances the performance of LLaVA-V1.6-7B by $2.47\%$, surpassing LoRA by $1.07\%$,  while requiring only $4.01\%$ and $4.35\%$ of LoRA's time cost and memory cost, respectively.

The main contributions of this paper can be summarized as follows:
\begin{itemize}
	\item We discuss the issues of high resource cost and parameter dependency associated inherent in PEFT methods.
	\item 
	We propose the EA, which independently complements the underfitted distributions of LMs on specific task.
	\item Extensive experiments indicate that EA facilitates a more democratized and scalable means of injecting task-specific knowledge into LMs.
\end{itemize}

\section{Related Work}
Existing PEFT methods can be broadly categorized into three types: additive fine-tuning, selective fine-tuning, and reparameterized fine-tuning \cite{han2024parameter}. Additive fine-tuning incorporates additional trainable components—such as adapters or prefix tokens—into the LM to capture task-specific knowledge without modifying the base model \cite{pfeiffer2020adapterfusion}. Selective fine-tuning focuses on fine-tuning a small subset of the LM's parameters, such as specific layers or parameter blocks, to reduce computational cost while preserving adaptability \cite{houlsby2019parameter,sung2021training}. Reparameterized fine-tuning modifies the parameterization of the model—for example, by using low-rank decomposition (e.g., LoRA)—to enable efficient learning with fewer trainable parameters \cite{dettmers2023qlora}. 

  \section{Method}
  
  The proposed Easy Adaptation (EA)  is a general framework that trains Specific Small Models (SSMs) on different tasks to complement the underfitted distributions of a Large Model (LM), enabling the LM to adapt to specific tasks. In this section, we will further elaborate on the implementation process of the proposed EA.

  \subsection{LM and SSM}
  LM, denoted as  $M_L$, with the vast parameter scale, are capable of fitting a wide range of data distributions, $D_L$. In contrast, SSM, denoted as $M_S$, due to the more limited parameter scale, can effectively fit only a narrower set of data distributions, $D_S$.
  
  Generally, the distribution $D_L$ exhibits a high degree of overlap with the data distributions of most tasks, which explains why LM perform well in zero-shot and few-shot settings. Building on this, fine-tuning is employed to incorporate task-specific knowledge into the LM, complementing for distributional gaps and thereby enhancing performance on specific tasks. When training a SSM for a specific task, the fitted distribution $D_S$, although significantly narrower than $D_L$, typically includes content absent from $D_L$ but essential for the task. 
  The formal expression of the above process is:
  \begin{equation}
  Scale(M_S)<<Scale(M_L), \quad Scale(D_S)<<Scale(D_L)
  \end{equation}
  where function $Scale(\cdot)$ measures the scale. 
  
  For an input $x$ of a specific task, when $x \in D_L$ and $x \notin D_S$, the input is not suitable for processing by the SSM, and the LM have a higher probability of producing the correct result for $x$:
  \begin{equation}
  \mathbb{P}(M_{L}(x)=y)>\mathbb{P}(M_{S}(x)=y )
  \end{equation}
  where $y$ is the ground truth.
  
  When $x \in D_S$ and $x \notin D_L$, the input $x$ originates from the distribution of a specific task learned by the SSM, which remains unknown to the LM. Thus: 
  \begin{equation}
  \mathbb{P}(M_{S}(x)=y)>\mathbb{P}(M_{L}(x)=y )
  \end{equation}
  When $x \in D_S \cap D_L$ or $x \notin D_S \cup D_L$, there is no significant difference between processing the input $x$ with the SSM or the LM.

  \subsection{Specific Layer: coarse-grained compensation for underfitted distribution of the LM}
  In order to decouple from the LMs' parameters and enable adaptation to specific tasks, we introduce the Specific Layer within the EA framework. As illustrated in Figure \ref{Figure: CA pipline} (a), Specific Layer consists of multiple SSMs trained on specific task datasets. Due to differences in model architecture, etc, each SSM may fit a distinct distribution $D_S$. By selecting the appropriate SSM for each input through a Router, the overall framework can cover a broader range of data distributions that LMs has not previously fitted. 
  
  Initially, we independently train $N$ SSMs, where each model $M_S^i$ is fitted to its unique distribution $D_S^i$. To assess whether a SSM is proficient at handling a given input $x$, we use the confidence output by the softmax layer $Softmax(\cdot)$ as the evaluation criterion \cite{niculescu2005predicting}:
  \begin{equation}
  C_S^i = max(Softmax(M_S^i(x))) >\tau, \quad x\in D_S^i
  \label{Eq: SL}
  \end{equation}
  where $\tau$ denotes the threshold that determines whether $M_S^i$ is proficient at handling the input $x$. 
  According to Eq.\ref{Eq: SL}, selecting different SSMs to process $x$ can significantly reduce the dependence on the LM, thereby reducing the cost associated with LMs.
  However, in Specific Layer, SSMs exhibit weak correlations with distributions that are poorly fitted by the LM, making it challenging to address the LM's capability deficiencies on specific tasks. Therefore, we further introduce the Augmented Layer.

  \subsection{Augmented Layer: targeted compensation for capability deficiencies of the LM}
  The SSMs within the Specific Layer only provide coarse-grained compensation and do not specifically target the underfitted distributions of the LM. To better complement for the deficiencies of LMs on specific tasks, we propose the Augmented Layer as an extension of the Specific Layer.

  As indicated by the blue arrows in Figure \ref{Figure: CA pipline} (a), the construction of the Augmented Layer begins with evaluating the SSMs in Specific Layer on the training set. Subsequently, the training data misclassified by the SSMs will be re-evaluated by the LM, thereby obtaining two subsets: the fitted dataset $X_{train}^{F}$ and the underfitted dataset $X_{train}^{U}$:
  \begin{align}
  \left\{
  \begin{aligned}
  &x \in X_{train}^{U},\quad\textnormal{where}\quad M_L(x) \neq y \quad\textnormal{and}\quad M_S(x) \neq y\\
  &x \in X_{train}^{F},\quad\textnormal{where} \quad M_L(x) = y\quad or\quad  M_S(x) = y
  \end{aligned}
  \right.
  \label{Eq: data}
  \end{align}
  After identifying the underfitted dataset $X_{train}^{U}$ of the Specific Layer and the LM by Eq.\ref{Eq: data}, we can train the Augmented Specific Small Models (ASSMs) to achieve targeted compensation:
  \begin{equation}
  M_{AS}=\arg \min _\theta \mathcal{L}\left(M_S(X_{train}^{U}), Y_{train}^{U}\right)
  \label{Eq: FT}
  \end{equation}
  where $\theta$ is the parameter set of the SSM $M_S$. Eq.\ref{Eq: FT} represents the fine-tuning process of $M_S$ with data $X_{train}^{U}$. With ASSM $M_{AS}$, we can further bridge the gaps in the fitted distributions of both SSMs and the LM, thereby enhancing the performance of the EA framework on specific tasks.

  \begin{table*}[t]
  	\caption{ The performance and resource costs of EA and PEFT methods across different tasks. }
  	\label{Table: Main}
  	\centering
  	\begin{tabularx}{\linewidth}{CMCCC}
  		\toprule
  		Task & Method & Time & Memory & Accuracy\\
  		\midrule
  		\multirow{11}{*}{NLI} 
  		& GLM-4-9B & \textemdash & \textemdash & 86.73\% \\
  		& GLM-4-9B-LoRA & 27:29:33 & 26390 & 90.62\% \\
  		& GLM-4-9B-QLoRA & 30:25:59 & 17392 & 90.68\% \\
  		& GLM-4-9B-Freeze & 9:23:53 & 35720 & 88.58\% \\
  		& EA (ours) & 8:30:21 & 4086 & 90.46\% \\
  		\cmidrule(r){2 - 5}
  		& Qwen2.5-32B & \textemdash & \textemdash & 89.64\% \\
  		& Qwen2.5-32B-LoRA & 52:15:36 & 76029 & 92.42\% \\
  		& Qwen2.5-32B-QLoRA & 108:23:31 & 29450 & 92.18\% \\
  		& Qwen2.5-32B-Freeze & 25:45:52 & 107836 & 91.08\% \\
  		& EA (ours) & 8:34:05 & 4086 & 91.06\% \\
  		\midrule
  		\multirow{11}{*}{SA}
  		& Llama-3-8B & \textemdash & \textemdash & 80.39\% \\
  		& Llama-3-8B-LoRA & 2:12:06 & 25626 & 81.21\% \\
  		& Llama-3-8B-QLoRA & 2:57:29 & 15098 & 81.09\% \\
  		& Llama-3-8B-Freeze & 0:47:34 & 30826 & 80.92\% \\
  		& EA  (ours) & 0:09:39 & 2962 & 81.53\% \\
  		\cmidrule(r){2 - 5}
  		& Qwen2.5-7B & \textemdash & \textemdash & 82.16\% \\
  		& Qwen2.5-7B-LoRA & 2:23:52 & 25560 & 82.96\% \\
  		& Qwen2.5-7B-QLoRA & 2:46:19 & 16172 & 82.72\% \\
  		& Qwen2.5-7B-Freeze & 0:44:52 & 32814 & 82.79\% \\
  		& EA (ours) & 0:09:18 & 2962 & 82.54\% \\
  		\midrule
  		\multirow{11}{*}{IC}
  		& Qwen2-VL-7B & \textemdash & \textemdash & 92.83\% \\
  		& Qwen2-VL-7B-LoRA & 10:36:37 & 17636 & 96.41\% \\
  		& Qwen2-VL-7B-QLoRA & 13:53:46 & 7402 & 96.45\% \\
  		& Qwen2-VL-7B-Freeze & 4:15:35 & 25524 & 95.08\% \\
  		& EA (ours) & 0:11:12 & 1058 & 95.56\% \\
  		\cmidrule(r){2 - 5}
  		& LLaVA-V1.6-7B & \textemdash & \textemdash & 93.57\% \\
  		& LLaVA-V1.6-7B-LoRA & 11:20:50 & 24304 & 94.97\% \\
  		& LLaVA-V1.6-7B-QLoRA & 13:03:26 & 14523 & 95.08\% \\
  		& LLaVA-V1.6-7B-Freeze & 6:34:24 & 27998 & 94.42\% \\
  		& EA (ours) & 0:11:55 & 1058 & 96.04\% \\
  		\bottomrule
  	\end{tabularx}
  \end{table*}

  \begin{table*}[t]
  	\caption{The performance and resource costs of EA and PEFT methods on SU that denotes summarization.}
  	\label{Table: SU}
  	\centering
  	\begin{tabularx}{\linewidth}{CMCCCCC}
  		\toprule
  		Task & Method & Time & Memory & R-1 & R-2 & R-L \\
  		\midrule
  		\multirow{10}{*}{SU}
  		& GLM-4-9B  & \textemdash & \textemdash & 34.90 & 12.47 & 24.39 \\
  		& GLM-4-9B-LoRA & 24:32:13 & 32749 & 40.25 & 16.34 & 26.06 \\
  		& GLM-4-9B-QLoRA & 28:10:34 & 22125 & 40.06 & 16.35 & 25.93 \\
  		& GLM-4-9B-Freeze & 9:06:55 & 36445 & 39.86 & 15.68 & 25.77 \\
  		& EA & 8:33:06 & 11834 & 40.17 & 15.91 & 26.10 \\
  		\cmidrule(r){2-7}
  		& Qwen2.5-14B & \textemdash & \textemdash & 36.82 & 13.27 & 26.14 \\
  		& Qwen2.5-14B-LoRA & 43:42:26 & 42760 & 42.82 & 18.25 & 27.50 \\
  		& Qwen2.5-14B-QLoRA & 50:33:47 & 20617 & 42.19 & 18.03 & 27.35 \\
  		& Qwen2.5-14B-Freeze & 13:07:37 & 48371 & 41.55 & 17.84 & 27.24 \\
  		& EA & 10:35:17 & 11834 & 41.32 & 18.86 & 27.47 \\
  		\bottomrule
  	\end{tabularx}
  \end{table*}

\begin{table*}[t]
	\caption{Results of Specific Layer within the EA framework. AL denotes Augmented Layer.}
	\label{Table: SL}
	\centering
	\begin{tabularx}{\linewidth}{CMCCC}
		\toprule
		Task & Model & Individual Accuracy& Proportion & Accuracy of AL\\
		\midrule
		\multirow{8}{*}{NLI} 
		& RoBERTa  & 85.51\% & 62.85\% & \multirow{4}{*}{90.06\%}  \\
		& DistilBERT  & 82.40\% & 5.21\% & \\
		& ALBERT  & 82.63\% & 2.75\% & \\
		& GLM-4-9B & 86.73\% & 29.18\% & \\
		\cmidrule(r){2 - 5}
		& RoBERTa & 85.51\% & 62.85\% & \multirow{4}{*}{90.38\%}  \\
		& DistilBERT & 82.40\% & 5.21\% & \\
		& ALBERT & 82.63\% & 2.75\% & \\
		& Qwen2.5-32B & 89.64\% & 29.18\% & \\
		\midrule
		\multirow{8}{*}{SA}
		& RoBERTa & 78.39\% & 71.78\% & \multirow{4}{*}{80.85\%}  \\
		& XLNet & 77.53\% & 3.83\% & \\
		& BERT & 76.26\% & 0.02\% & \\
		& Llama-3-8B & 80.39\% & 24.37\% & \\
		\cmidrule(r){2 - 5}
		& RoBERTa & 78.39\% & 41.17\% & \multirow{4}{*}{82.29\%}  \\
		& XLNet & 77.53\% & 0.04\% & \\
		& BERT & 76.26\% & 0.1\% & \\
		& Qwen2.5-7B & 82.16\% & 58.69\% & \\
		\midrule
		\multirow{8}{*}{IC}
		& MobileNet V2  & 90.53\% & 41.84\% & \multirow{4}{*}{94.93\%}  \\
		& ResNet-34  & 89.87\% & 24.44\% & \\
		& SqueezeNet & 82.99\% & 7.58\% & \\
		& Qwen2-VL-7B & 92.83\% & 26.25\% & \\
		\cmidrule(r){2 - 5}
		& MobileNet V2 & 90.53\% & 41.84\% & \multirow{4}{*}{94.67\%}  \\
		& ResNet-34 & 89.87\% & 24.44\% & \\
		& SqueezeNet & 82.99\% & 7.58\% & \\
		& LLaVA-V1.6-7B & 93.57\% & 26.25\% & \\
		\bottomrule
	\end{tabularx}
\end{table*}

\begin{table*}[t]
	\caption{Comparison of results with and without targeted compensation. }
	\label{Table: targeted compensation}
	\centering
	\begin{tabularx}{\linewidth}{CCCCCC}
		\toprule
		Task & Model & Method & Time & Memory & Accuracy \\
		\midrule
		\multirow{2}{*}{NLI} 
		& \multirow{2}{*}{GLM-4-9B} 
		& EA(Full) & 12:54:35 & 4086 & 90.40\% \\
		&  & EA      & 8:30:21  & 4086 & 90.46\% \\
		\midrule
		\multirow{2}{*}{SA}
		& \multirow{2}{*}{Qwen2.5-7B}
		& EA(Full) & 0:20:57 & 2962 & 82.70\% \\
		&  & EA      & 0:09:18 & 2962 & 82.54\% \\
		\midrule
		\multirow{2}{*}{IC}
		& \multirow{2}{*}{Qwen2-VL-7B}
		& EA(Full) & 7:56:04 & 1058 & 95.64\% \\
		&  & EA      & 0:11:12 & 1058 & 95.56\% \\
		\bottomrule
	\end{tabularx}
\end{table*}

  \subsection{Router}
  The Specific Layer and Augmented Layer provide the foundational capabilities for the LM to adapt to specific tasks. However, during inference, when a new input is received, a routing mechanism is still needed to determine the input should be processed by which model. 
  According to the prior work \cite{niculescu2005predicting}, confidence $C_S^i$ in Eq.\ref{Eq: SL} can serve as a criterion for evaluating whether the model is capable of handling the input. To substitute SSMs and ASSMs for LMs in fitting data during task-specific fine-tuning, thus allowing LMs to adapt to new tasks, we propose the Router. As presented in Figure \ref{Figure: CA pipline} (b), the Router is implemented based on the model performance, arranging the execution order of models and returning results when certain conditions are met, thereby enabling multi-stage processing of inputs belonging to different distributions. Specifically, for Router 1 in Figure \ref{Figure: CA pipline} (a), after training $N$ SSMs, we evaluate SSMs on the validation set $X_{val}$ by: 
  \begin{equation}
  { P^i }=\frac{1}{J} \sum_{i=1}^J  \mathbb{I} \left(M_S^i\left(x_i\right)=y_i\right), x_i \in X_{val}
  \label{Eq: performance}
  \end{equation}
  where $\mathbb{I}$ is indicator. With performance evaluated by Eq.\ref{Eq: performance}, we can rank SSMs in descending order:
  \begin{equation}
  M_S^{Index(P_1)}\geq M_S^{Index(P_2)} \geq \cdots \geq M_S^{Index(P_N)}
  \label{Eq: Router}
  \end{equation}
  where $P_1$ corresponds to the best SSM, while that of $P_N$ corresponds to the worst one, $Index(\cdot)$ is the index function that can return the model id.
  During inference, input $x$ will first be processed by $M_S^{Index(P_1)}$. If confidence $C_S^{Index(P_1)}\geq\tau_1$, the processing of the input data $x$ will terminate. Conversely, if  $C_S^{Index(P_1)}<\tau_1$, the input $x$ will be passed to the next SSM, and the above procedure will be repeated. When the input $x$ does not belong to the distribution fitted by any SSM, it will ultimately be handled by the LM. The mechanism of Router 2 in Figure \ref{Figure: CA pipline} (a) is identical to that of Router 1, with the sole distinction that the Augmented Layer is activated only when LM cannot handle the input. 
  The whole process of EA is summarized in Appendix A. 

  \section{Experiments}
  In our experiments, we aim to 
  (1) evaluate whether EA can enable the LM to adapt to specific tasks, while incurring minimal local resource costs compared to PEFT methods.
  (2) evaluate the effectiveness of Specific Layer within EA to achieve coarse-grained compensation for deficient capabilities of the LM,
  (3) evaluate the effectiveness of Augmented Layer within EA to achieve targeted compensation for capability deficiencies of the LM,
  (4) evaluate whether EA can eliminate dependence on the LM’s parameters, thereby achieving task adaptation based on APIs.
  The code and data for the proposed method are provided for research purposes \footnote{Code is included in the supplemental material and will be released upon the paper acceptance.}.
Experiments are conducted on XNLI \cite{conneau2018xnli} for the Natural Language Inference (NLI) task, Yelp Reviews \cite{asghar2016yelp} for Sentiment Analysis (SA), CIFAR-10 \cite{krizhevsky2009learning} for Image Classification (IC), and CNN/DailyMail \cite{nallapati2016abstractive} for Summarization.
  Unless otherwise specified, the small model used by EA is RoBERTa \cite{liu2019roberta}, MobileNet V2 \cite{sandler2018mobilenetv2} and T5 \cite{raffel2020exploring}. 
  Moreover, in the following experiments, the reported memory cost reflects only the memory required for training the local SSMs, while the time cost includes both the cloud-based LM inference time and the local SSMs training time.
  More details about the datasets and experimental settings please refer to the Appendix B. 
  
  \begin{table*}[t]
  	\caption{EA for closed-source LMs.}
  	\label{Table: API}
  	\centering
  	\begin{tabularx}{\linewidth}{CMC|CMC}
  		\toprule
  		Task & Method & Acc & Task & Method & Acc\\
  		\midrule
  		\multirow{9}{*}{NLI}
  		& \small{GLM-4-Plus} & 90.56\% & \multirow{9}{*}{IC} & \small{GLM-4V-Plus} & 95.01\% \\
  		& \small{EA (ours)} & 91.20\% & & \small{EA (ours)} & 96.52\% \\
  		\cmidrule(r){2 - 3}
  		\cmidrule(r){5 - 6}
  		& \small{Doubao-1.5-pro-32k} & 89.98\% & & \small{Doubao-1.5-vision-pro} & 93.38\% \\
  		& \small{EA (ours)} & 91.14\% & & \small{EA (ours)} & 95.41\% \\
  		\cmidrule(r){2 - 3}
  		\cmidrule(r){5 - 6}
  		& \small{Qwen-Max} & 90.10\% & & \small{Qwen-VL-Max} & 95.41\%\\
  		& \small{EA (ours)} & 90.92\% & & \small{EA (ours)} & 96.19\%\\
  		\cmidrule(r){2 - 3}
  		\cmidrule(r){5 - 6}
  		& \small{Hunyuan-Turbos} & 88.62\% & & \small{Hunyuan-Turbos-vision} & 94.71\%\\
  		& \small{EA (ours)} & 90.18\% & & \small{EA (ours)} & 96.63\%\\
  		\bottomrule
  	\end{tabularx}
  \end{table*}

 \begin{table*}[!t]
	\caption{The impact of LM adaptation to specific tasks on different data types.}
	\label{Table: When CA performs better}
	\centering
	\begin{tabularx}{\linewidth}{MCCCCCCC}
		\toprule
		Method & Time & Memory& Overall & Region & Head & Medium & Tail\\
		\midrule
		Qwen2-VL-7B & \textemdash & 33108 & 92.83\% & 92.32\% & 93.61\% & 89.42\% & 93.92\% \\
		Qwen2-VL-7B-LoRA & 10:36:37 & 17636 & 96.41\% & 94.91\% & 97.35\% & 93.46\% & 93.92\% \\
		Qwen2-VL-7B-QLoRA & 13:53:46 & 7402 & 96.45\% & 95.09\% & 97.40\% & 93.27\%& 94.59\% \\
		Qwen2-VL-7B-Freeze & 4:15:35 & 25524 & 95.08\% & 90.74\% & 97.94\% & 85.77\% & 88.51\% \\
		EA (ours)& 0:11:12 & 1058 & 95.56\% & 92.21\% & 97.45\% & 90.00\% & 89.19\% \\
		\midrule
		LLaVA-V1.6-7B & \textemdash & 31628 & 93.57\% & 89.82\% & 94.35\% & 94.04\% & 81.08\% \\
		LLaVA-V1.6-7B-LoRA & 11:20:50 & 24304 & 94.97\% & 86.77\% & 97.35\% & 92.69\% & 70.27\% \\
		LLaVA-V1.6-7B-QLoRA & 13:03:26 & 14523 & 95.08\% & 86.91\% & 97.40\% & 93.08\% & 70.27\% \\
		LLaVA-V1.6-7B-Freeze & 6:34:24 & 27998 & 94.42\% & 86.62\% & 96.91\% & 91.15\% & 71.81\% \\
		EA (ours)& 0:11:55 & 1058 & 96.04\% & 91.94\% & 97.59\% & 93.08\% & 85.14\% \\
		\bottomrule
	\end{tabularx}
\end{table*}

  \subsection{Comparison of EA with PEFT methods}
  Currently, the most popular PEFT methods, such as LoRA \cite{hu2022lora}, QLoRA \cite{dettmers2023qlora}, and Freeze, have significantly reduced the resources required for fine-tuning LMs.
  However, for the majority of individuals using relatively low-cost devices, the resource requirements of PEFT remain prohibitively high. In particular, many users significantly reduce the batch size to lower memory usage, which in turn leads to substantial increases in time cost.
  To alleviate the above issue, we propose EA that injects specific knowledge by training SSMs, and as a result, the resource cost is comparable to model training before the era of LMs. In Table \ref{Table: Main}, we compare the effectiveness of EA, LoRA, QLoRA, and Freeze in adapting LMs to specific classification tasks across different datasets.
  
  As shown in the Table \ref{Table: Main}, EA achieves results very close to the best-performing PEFT methods in most cases, while significantly reducing resource cost. For example, in the NLI task, QLoRA improves the performance of GLM-4-9B from $86.73\%$ to $90.68\%$. EA achieves a comparable result of $90.46\%$, while consuming only about one-quarter of QLoRA's memory and one-third of its training time. In addition, in the IC task, EA boosted the performance of LLaVA-V1.6-7B from $93.57\%$ to $96.04\%$, outperforming other PEFT methods, while its time and memory cost are only one-thirteenth and one-twenty-eighth, respectively, of those of the best-performing PEFT method. 
Furthermore,  we present the results of EA on the generation task in Table \ref{Table: SU}. As can be observed, compared with the classification tasks reported in Table \ref{Table: Main}, the generation task incurs relatively higher time and memory cost due to the larger local model (T5). Nevertheless, the cost remains only about one-third to one-quarter of that required by LoRA.
  
  The aforementioned results indicate that EA can achieve performance comparable to existing PEFT methods while more efficiently conserving computational resources. By invoking the original LMs via APIs, all EA-related results in Table \ref{Table: Main} can be rapidly obtained with a RTX2050 (4 GB).

  \subsection{Ablation: The effectiveness of Specific Layer for coarse-grained compensation}
  
  The Specific Layer in the EA framework offers coarse-grained capability compensation to assist the LM in adapting to specific tasks. Although the SSMs in Specific Layer cannot provide targeted compensation for underfitted distributions, they route a portion of the inputs away from the LM via the Router. This substantially reduces the LM’s invocation frequency, thereby conserving computational resources.
  
  In Table \ref{Table: SL}, we present the model invocation proportions within Specific Layers, as well as their performance. It can be observed that, compared to the original LM, the Specific Layer achieves comparable results through model collaboration, while significantly reducing the invocation rate of the LM. For instance, for the Llama-3-8B of SA task, Specific Layer achieves a result of  $80.85\%$, slightly higher than the LM's $80.39\%$, while the invocation rate of the LM is only $24.37\%$. This indicates that the time required to complete task-specific inference can be significantly reduced, as the majority of inputs will be processed by the smaller, faster SSMs. In addition, the results in Table \ref{Table: SL} show significant variation in the invocation proportions of SSMs within the Specific Layer. Taking the previously mentioned SA task as an example, in the Router module that routes inputs based on performance ranking, RoBERTa is invoked first and handles $71.78\%$ of the inputs. XLNet follows, processing only $3.83\%$ of the inputs, while BERT, invoked last, handles merely $0.02\%$. These results suggest that lower-ranked SSMs in the Specific Layer may contribute minimally.

  \subsection{Ablation: The effectiveness of Augmented Layer for targeted compensation}
  EA collects underfitted data from the Specific Layer and the LM on the training dataset, and further trains ASSMs to compensate for the model’s capability deficiencies. This section mainly validates the effectiveness of targeted compensation within the EA framework. The corresponding results are summarized in Table \ref{Table: targeted compensation}, where EA(Full) refers to the Augmented Layer are fine-tuned on all data where the LM makes errors.

  As shown in the Table \ref{Table: targeted compensation}, the performance of targeted compensation (i.e., EA) is comparable to that of fine-tuning with all misclassified data (i.e., EA(Full)). Meanwhile, since EA collects underfitted data that is filtered by both the SSMs and the LM, the resulting dataset is smaller in size but more targeted in content.
  As a result, EA with targeted compensation offers a clear advantage in training efficiency. For instance, for NLI and SA, the training time of EA is about one-half to two-thirds of that of EA(Full), while for IC, the training time of EA is about $2.31\%$ of that of EA(Full). Additional experiments with different LLMs are provided in Appendix C.

    \subsection{EA for closed-source LMs}
  As competition among LMs intensifies, an increasing number of companies are adopting closed-source policies and providing services to users via APIs. 
  As PEFT methods heavily rely on parameters update, making it impossible to adapt closed-source LMs to specific tasks. On the other hand, uploading datasets for cloud-based fine-tuning incurs significant costs. 
  We demonstrate the EA framework can enable closed-source LMs to adapt to specific tasks, and related results are presented in Table \ref{Table: API}.
  Obviously, the EA framework yields notable improvements for various closed-source LMs on both NLI and IC tasks. In particular, for the IC task, EA improves the performance of  Doubao-1.5-pro-32k by $2.03\%$. All results in Table \ref{Table: API} clearly demonstrate that the EA framework can effectively adapt LMs to specific tasks without requiring access to model parameters, thereby offering broader applicability across diverse scenarios.

  \subsection{When EA performs better}
  In Table \ref{Table: When CA performs better}, we present a fine-grained analysis of the impact of different methods on the LM by partitioning the CIFAR-10 based on the number of samples in each category.
  For Qwen2VL-7B, QLoRA improves the overall accuracy of the LM from $92.83\%$ to $96.45\%$, with its fine-tuning process primarily enhancing the performance on head and middle data, resulting in accuracy gains of $3.79\%$ and $3.85\%$, respectively. In contrast, EA mainly addresses the LM’s underfitting on head data, while its performance on tail data even slightly declines. This leads to EA's relatively weaker performance for Qwen2-VL-7B.
  As for LLaVA-V1.6-7B, it demonstrates strong performance on head and middle data, which contributes to its overall better results compared to Qwen2-VL-7B. However, due to its excellent fitting on head data and the large proportion of head samples, the model becomes overfitted to head data during fine-tuning. Such overfitting negatively affects the performance of PEFT methods on middle and tail data. In contrast, EA mitigates this issue by training targeted small models to fit specific task distributions, making it less susceptible to overfitting.
  Overall, when the original LM is more prone to overfitting the specific task distribution, the EA framework tends to exhibit greater performance advantages.
  
    More experiments about hyperparameter analysis and ablation study please refer to the Appendix D and E.

  \section{Conclusion}
  Parameter Efficient Fine-Tuning (PEFT) methods are crucial for the widespread adoption of Large Models (LM). Although PEFT methods, such as LoRA, have significantly reduced the difficulty of fine-tuning LMs, they still face challenges in extreme scenarios, including high resource cost and parameter dependency. To address these issues, we propose training Specific Small Models (SSMs) to complement the underfitted distributions of LMs on specific tasks, which enables fast, lightweight, and parameter-free adaptation of LMs. Extensive experiments validate the effectiveness of the proposed framework, particularly in resource-constrained environments.

\bibliographystyle{named}
\bibliography{ijcai26}

\begin{thebibliography}{}

\bibitem[\protect\citeauthoryear{Ansell \bgroup \em et al.\egroup
  }{2021}]{ansell2021composable}
Alan Ansell, Edoardo~Maria Ponti, Anna Korhonen, and Ivan Vuli{\'c}.
\newblock Composable sparse fine-tuning for cross-lingual transfer.
\newblock {\em arXiv preprint arXiv:2110.07560}, 2021.

\bibitem[\protect\citeauthoryear{Asghar}{2016}]{asghar2016yelp}
Nabiha Asghar.
\newblock Yelp dataset challenge: Review rating prediction.
\newblock {\em arXiv preprint arXiv:1605.05362}, 2016.

\bibitem[\protect\citeauthoryear{Bahri \bgroup \em et al.\egroup
  }{2024}]{bahri2024explaining}
Yasaman Bahri, Ethan Dyer, Jared Kaplan, Jaehoon Lee, and Utkarsh Sharma.
\newblock Explaining neural scaling laws.
\newblock {\em Proceedings of the National Academy of Sciences},
  121(27):e2311878121, 2024.

\bibitem[\protect\citeauthoryear{Borzunov \bgroup \em et al.\egroup
  }{2023}]{borzunov2023distributed}
Alexander Borzunov, Max Ryabinin, Artem Chumachenko, Dmitry Baranchuk, Tim
  Dettmers, Younes Belkada, Pavel Samygin, and Colin~A Raffel.
\newblock Distributed inference and fine-tuning of large language models over
  the internet.
\newblock {\em Advances in neural information processing systems},
  36:12312--12331, 2023.

\bibitem[\protect\citeauthoryear{Chen \bgroup \em et al.\egroup
  }{2024}]{chen2024data}
Dong Chen, Yueting Zhuang, Shuo Zhang, Jinfeng Liu, Su~Dong, and Siliang Tang.
\newblock Data shunt: Collaboration of small and large models for lower costs
  and better performance.
\newblock In {\em Proceedings of the AAAI Conference on Artificial
  Intelligence}, volume~38, pages 11249--11257, 2024.

\bibitem[\protect\citeauthoryear{Chen \bgroup \em et al.\egroup
  }{2025a}]{chen2025improving}
Dong Chen, Fei Gao, Shuo Zhang, Yueting Zhuang, Siliang Tang, Qidong Liu, Hua
  Wang, Xin Yang, and Mingliang Xu.
\newblock Improving large models with small models: Lower costs and better
  performance.
\newblock {\em Neural Networks}, page 108276, 2025.

\bibitem[\protect\citeauthoryear{Chen \bgroup \em et al.\egroup
  }{2025b}]{chen2025logic}
Dong Chen, Shilin Zhang, Fei Gao, Yueting Zhuang, Siliang Tang, Qidong Liu, and
  Mingliang Xu.
\newblock Logic distillation: learning from code function by function for
  decision-making tasks.
\newblock In {\em Proceedings of the Thirty-Fourth International Joint
  Conference on Artificial Intelligence}, pages 7338--7346, 2025.

\bibitem[\protect\citeauthoryear{Conneau \bgroup \em et al.\egroup
  }{2018}]{conneau2018xnli}
Alexis Conneau, Ruty Rinott, Guillaume Lample, Holger Schwenk, Ves Stoyanov,
  Adina Williams, and Samuel~R Bowman.
\newblock Xnli: Evaluating cross-lingual sentence representations.
\newblock In {\em 2018 Conference on Empirical Methods in Natural Language
  Processing, EMNLP 2018}, pages 2475--2485. Association for Computational
  Linguistics, 2018.

\bibitem[\protect\citeauthoryear{Das \bgroup \em et al.\egroup
  }{2023}]{das2023unified}
Sarkar Snigdha~Sarathi Das, Ranran~Haoran Zhang, Peng Shi, Wenpeng Yin, and Rui
  Zhang.
\newblock Unified low-resource sequence labeling by sample-aware dynamic sparse
  finetuning.
\newblock {\em arXiv preprint arXiv:2311.03748}, 2023.

\bibitem[\protect\citeauthoryear{Dettmers \bgroup \em et al.\egroup
  }{2023}]{dettmers2023qlora}
Tim Dettmers, Artidoro Pagnoni, Ari Holtzman, and Luke Zettlemoyer.
\newblock Qlora: Efficient finetuning of quantized llms.
\newblock {\em Advances in neural information processing systems},
  36:10088--10115, 2023.

\bibitem[\protect\citeauthoryear{Gheini \bgroup \em et al.\egroup
  }{2021}]{gheini2021cross}
Mozhdeh Gheini, Xiang Ren, and Jonathan May.
\newblock Cross-attention is all you need: Adapting pretrained transformers for
  machine translation.
\newblock {\em arXiv preprint arXiv:2104.08771}, 2021.

\bibitem[\protect\citeauthoryear{Han \bgroup \em et al.\egroup
  }{2024}]{han2024parameter}
Zeyu Han, Chao Gao, Jinyang Liu, Jeff Zhang, and Sai~Qian Zhang.
\newblock Parameter-efficient fine-tuning for large models: A comprehensive
  survey.
\newblock {\em arXiv preprint arXiv:2403.14608}, 2024.

\bibitem[\protect\citeauthoryear{Houlsby \bgroup \em et al.\egroup
  }{2019}]{houlsby2019parameter}
Neil Houlsby, Andrei Giurgiu, Stanislaw Jastrzebski, Bruna Morrone, Quentin
  De~Laroussilhe, Andrea Gesmundo, Mona Attariyan, and Sylvain Gelly.
\newblock Parameter-efficient transfer learning for nlp.
\newblock In {\em International conference on machine learning}, pages
  2790--2799. PMLR, 2019.

\bibitem[\protect\citeauthoryear{Hu \bgroup \em et al.\egroup
  }{2022}]{hu2022lora}
Edward~J Hu, Yelong Shen, Phillip Wallis, Zeyuan Allen-Zhu, Yuanzhi Li, Shean
  Wang, Lu~Wang, Weizhu Chen, et~al.
\newblock Lora: Low-rank adaptation of large language models.
\newblock {\em ICLR}, 1(2):3, 2022.

\bibitem[\protect\citeauthoryear{Kaplan \bgroup \em et al.\egroup
  }{2020}]{kaplan2020scaling}
Jared Kaplan, Sam McCandlish, Tom Henighan, Tom~B Brown, Benjamin Chess, Rewon
  Child, Scott Gray, Alec Radford, Jeffrey Wu, and Dario Amodei.
\newblock Scaling laws for neural language models.
\newblock {\em arXiv preprint arXiv:2001.08361}, 2020.

\bibitem[\protect\citeauthoryear{Krizhevsky \bgroup \em et al.\egroup
  }{2009}]{krizhevsky2009learning}
Alex Krizhevsky, Geoffrey Hinton, et~al.
\newblock Learning multiple layers of features from tiny images.
\newblock 2009.

\bibitem[\protect\citeauthoryear{Lei \bgroup \em et al.\egroup
  }{2023}]{lei2023conditional}
Tao Lei, Junwen Bai, Siddhartha Brahma, Joshua Ainslie, Kenton Lee, Yanqi Zhou,
  Nan Du, Vincent Zhao, Yuexin Wu, Bo~Li, et~al.
\newblock Conditional adapters: Parameter-efficient transfer learning with fast
  inference.
\newblock {\em Advances in Neural Information Processing Systems},
  36:8152--8172, 2023.

\bibitem[\protect\citeauthoryear{Li and Liang}{2021}]{li2021prefix}
Xiang~Lisa Li and Percy Liang.
\newblock Prefix-tuning: Optimizing continuous prompts for generation.
\newblock {\em arXiv preprint arXiv:2101.00190}, 2021.

\bibitem[\protect\citeauthoryear{Liu \bgroup \em et al.\egroup
  }{2019}]{liu2019roberta}
Yinhan Liu, Myle Ott, Naman Goyal, Jingfei Du, Mandar Joshi, Danqi Chen, Omer
  Levy, Mike Lewis, Luke Zettlemoyer, and Veselin Stoyanov.
\newblock Roberta: A robustly optimized bert pretraining approach.
\newblock {\em arXiv preprint arXiv:1907.11692}, 2019.

\bibitem[\protect\citeauthoryear{Nallapati \bgroup \em et al.\egroup
  }{2016}]{nallapati2016abstractive}
Ramesh Nallapati, Bowen Zhou, Cicero Dos~Santos, {\c{C}}a{\u{g}}lar
  Gul{\c{c}}ehre, and Bing Xiang.
\newblock Abstractive text summarization using sequence-to-sequence rnns and
  beyond.
\newblock In {\em Proceedings of the 20th SIGNLL conference on computational
  natural language learning}, pages 280--290, 2016.

\bibitem[\protect\citeauthoryear{Niculescu-Mizil and
  Caruana}{2005}]{niculescu2005predicting}
Alexandru Niculescu-Mizil and Rich Caruana.
\newblock Predicting good probabilities with supervised learning.
\newblock In {\em Proceedings of the 22nd international conference on Machine
  learning}, pages 625--632, 2005.

\bibitem[\protect\citeauthoryear{Pfeiffer \bgroup \em et al.\egroup
  }{2020}]{pfeiffer2020adapterfusion}
Jonas Pfeiffer, Aishwarya Kamath, Andreas R{\"u}ckl{\'e}, Kyunghyun Cho, and
  Iryna Gurevych.
\newblock Adapterfusion: Non-destructive task composition for transfer
  learning.
\newblock {\em arXiv preprint arXiv:2005.00247}, 2020.

\bibitem[\protect\citeauthoryear{Raffel \bgroup \em et al.\egroup
  }{2020}]{raffel2020exploring}
Colin Raffel, Noam Shazeer, Adam Roberts, Katherine Lee, Sharan Narang, Michael
  Matena, Yanqi Zhou, Wei Li, and Peter~J Liu.
\newblock Exploring the limits of transfer learning with a unified text-to-text
  transformer.
\newblock {\em Journal of machine learning research}, 21(140):1--67, 2020.

\bibitem[\protect\citeauthoryear{Sandler \bgroup \em et al.\egroup
  }{2018}]{sandler2018mobilenetv2}
Mark Sandler, Andrew Howard, Menglong Zhu, Andrey Zhmoginov, and Liang-Chieh
  Chen.
\newblock Mobilenetv2: Inverted residuals and linear bottlenecks.
\newblock In {\em Proceedings of the IEEE conference on computer vision and
  pattern recognition}, pages 4510--4520, 2018.

\bibitem[\protect\citeauthoryear{Sung \bgroup \em et al.\egroup
  }{2021}]{sung2021training}
Yi-Lin Sung, Varun Nair, and Colin~A Raffel.
\newblock Training neural networks with fixed sparse masks.
\newblock {\em Advances in Neural Information Processing Systems},
  34:24193--24205, 2021.

\bibitem[\protect\citeauthoryear{Touvron \bgroup \em et al.\egroup
  }{2023}]{touvron2023llama}
Hugo Touvron, Thibaut Lavril, Gautier Izacard, Xavier Martinet, Marie-Anne
  Lachaux, Timoth{\'e}e Lacroix, Baptiste Rozi{\`e}re, Naman Goyal, Eric
  Hambro, Faisal Azhar, et~al.
\newblock Llama: Open and efficient foundation language models.
\newblock {\em arXiv preprint arXiv:2302.13971}, 2023.

\bibitem[\protect\citeauthoryear{Valipour \bgroup \em et al.\egroup
  }{2022}]{valipour2022dylora}
Mojtaba Valipour, Mehdi Rezagholizadeh, Ivan Kobyzev, and Ali Ghodsi.
\newblock Dylora: Parameter efficient tuning of pre-trained models using
  dynamic search-free low-rank adaptation.
\newblock {\em arXiv preprint arXiv:2210.07558}, 2022.

\bibitem[\protect\citeauthoryear{Zhang \bgroup \em et al.\egroup
  }{2023}]{zhang2023adalora}
Qingru Zhang, Minshuo Chen, Alexander Bukharin, Nikos Karampatziakis, Pengcheng
  He, Yu~Cheng, Weizhu Chen, and Tuo Zhao.
\newblock Adalora: Adaptive budget allocation for parameter-efficient
  fine-tuning.
\newblock {\em arXiv preprint arXiv:2303.10512}, 2023.

\bibitem[\protect\citeauthoryear{Zheng \bgroup \em et al.\egroup
  }{2025}]{zheng2025large}
Yanxin Zheng, Wensheng Gan, Zefeng Chen, Zhenlian Qi, Qian Liang, and Philip~S
  Yu.
\newblock Large language models for medicine: a survey.
\newblock {\em International Journal of Machine Learning and Cybernetics},
  16(2):1015--1040, 2025.

\end{thebibliography}

\end{document}